\documentclass{article}

\usepackage[preprint]{neurips_2026}
\usepackage[utf8]{inputenc} % allow utf-8 input
\usepackage[T1]{fontenc}    % use 8-bit T1 fonts
\usepackage{hyperref}       % hyperlinks
\usepackage{url}            % simple URL typesetting
\usepackage{booktabs}       % professional-quality tables
\usepackage{amsfonts}       % blackboard math symbols
\usepackage{amsmath}        % for \text and other math features
\usepackage{nicefrac}       % compact symbols for 1/2, etc.
\usepackage{microtype}      % microtypography
\usepackage{graphbox}      % for \graph command
\usepackage{xcolor}         % colors

\usepackage{graphicx}
\usepackage{tabularx}   % For auto-wrapping table columns
\usepackage{caption}

\usepackage{float}

\usepackage{caption}
\usepackage{titlesec}
\titlespacing*{\subsection}
{0pt}{6pt}{3pt}
\titlespacing*{\section}
{0pt}{6pt}{3pt}

\title{Robust Checkpoint Selection for Multimodal LLMs via Multi-Stage Evaluation and Stability-Aware Ranking}

\author{
  Qinwu Xu$^{1}$ \quad Zhuoheng Li$^{1}$ \quad Jessie Salas$^{1*}$ \\
  $^{1}$Meta AI \\
  $^{*}$Currently Independent Researcher
}

\begin{document}

\maketitle

\begin{abstract}
Selecting a final checkpoint for multimodal large language models
(MLLMs) is challenging when late-stage candidates are closely matched
and downstream evaluation signals are noisy. Small observed differences
can be comparable to variability introduced by finite evaluation
samples, LLM-based judges, and ambiguous multimodal evidence, while
validation loss may not identify the checkpoint preferred by downstream
evaluation. We formulate late-stage checkpoint selection as a
stability-aware decision problem under evaluation uncertainty and
propose a progressive framework combining pointwise filtering, listwise
ranking, and pairwise refinement. Repeated evaluation-set subsampling
is used to characterize ranking stability, while percentile-based
aggregation accounts for lower- and upper-tail behavior. Experiments
show that multimodal data evaluability is critical: quality-aware
curation of OCR-heavy inputs reduces ranking flip rate from 32.5\% to
11.2\% and increases inter-run agreement from 0.61 to 0.84. We further
observe divergence between validation-loss progression and downstream
checkpoint preference in two independent MLLM settings. An additional
public Qwen2.5-VL-7B reproduction across 11 checkpoints shows tightly
clustered pointwise scores and frequently tie-dominated final pairwise
comparisons, while repeated evaluation most often selects an
intermediate rather than the final checkpoint. These results suggest
that reliable MLLM checkpoint selection should quantify and preserve
evaluation uncertainty rather than force decisions from small
differences in a single metric.
\end{abstract}

\section{Introduction}

Selecting a final checkpoint is a consequential but often under-specified
step in multimodal large language model (MLLM) post-training.
Conventional practice commonly relies on validation loss, benchmark
performance, or automated evaluation to determine when training has
converged and which checkpoint should be deployed (Prechelt, 1998;
Prechelt, 2002). Meanwhile, LLM-based evaluators have become increasingly
common for open-ended model evaluation, using pointwise scoring, pairwise
comparison, or rubric-guided judgment (Zheng et al., 2023; Liu et al.,
2023; Kim et al., 2023). These approaches provide scalable evaluation
signals, but selecting among closely matched late-stage checkpoints
presents a different challenge: small observed performance differences
may be comparable to the variability introduced by evaluation-set
sampling, autoregressive generation, judge behavior, and multimodal
input quality.

This problem is particularly challenging for MLLMs evaluated on real-world visual inputs. Existing studies have shown substantial gaps between performance on conventional multimodal benchmarks and more realistic evaluation settings, including OCR-intensive and real-world visual reasoning tasks (Zhang et al., 2024; Fu et al., 2025).Images captured from wearable or mobile devices may contain small or partially visible text, blur, occlusion, unusual viewpoints, and ambiguous visual evidence (Xu et al., 2026; Xu, 2026). In such settings, an evaluation sample may itself be difficult to judge reliably. Consequently, a small difference in aggregate score does not necessarily imply a meaningful difference between checkpoints.

As illustrated in Figure~\ref{fig:receipt_candidate_eval}, this difficulty can arise
even on seemingly simple multimodal examples. Pointwise evaluation may
assign noticeably different scores to semantically equivalent correct
responses while assigning non-trivial scores to incorrect responses.
When checkpoint differences are already small, such score variation can
obscure the underlying ordering. Relative comparison provides a more
discriminative signal, but it does not by itself establish whether the
resulting ranking is stable under evaluation variability.

% \begin{figure}[t]
%     \centering
%     \includegraphics[width=\linewidth]{receipt_total_cost_figure1_v2.png}
%     \caption{\textbf{Motivating example for late-stage checkpoint selection.
%     Pointwise evaluation produces inconsistent or poorly separated
%     scores across closely competing responses, whereas relative
%     comparison provides a more discriminative ordering. The receipt
%     image was captured by the authors.}}
%     \label{fig:pscore2}
% \end{figure}

\begin{figure*}[htbp]
  \centering
  % --- LEFT SIDE: Question + Image ---
  \begin{minipage}[t]{0.24\textwidth}
    \vspace{0pt} % Aligns top of minipage
    \textbf{Question:} \textit{What is the total amount?}
    
    \vspace{0.5em}
    \includegraphics[width=\textwidth]{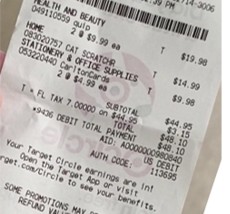} % Replace with your image file
  \end{minipage}%
  \hfill
  % --- RIGHT SIDE: Evaluation Table ---
  \begin{minipage}[t]{0.74\textwidth}
    \vspace{0pt} % Aligns top of minipage
    \scriptsize
    \setlength{\tabcolsep}{3pt} % Shrinks space between columns to maximize text space
    
    % Custom column width distribution: Response gets 0.77x weight, Rationale gets 1.23x weight
    \begin{tabularx}{\textwidth}{c >{\hsize=0.77\hsize}X c c c >{\hsize=1.23\hsize}X}
      \toprule
      \textbf{Cand.} & \textbf{Response} & \textbf{P.Score} & \textbf{P.Rank} & \textbf{L.Rank} & \textbf{Rationale for Listwise Rank} \\
      \midrule
      \textbf{C1} & The total amount paid is \$48.10. & 0.92 & \#2 & \#1 & Optimal response. It is direct, accurate, and naturally identifies the final amount paid (\$48.10). \\
      \midrule
      \textbf{C2} & The receipt shows a total of \$48.10... & 0.94 & \#1 & \#3 & Highly accurate. Clear and includes units, though slightly more wordy than C1. \\
      \midrule
      \textbf{C3} & The total amount on the receipt is \$48.10 dollars. & 0.91 & \#3 & \#2 & Accurate. Correct total, but the trailing ellipsis makes the response feel less polished. \\
      \midrule
      \textbf{C4} & Adding all items... gives \$48.10 & 0.89 & \#4 & \#4 & Technically correct. However, the rationale of ``Adding all items...'' is unnecessary as the total is already printed. \\
      \midrule
      \textbf{C5} & The total amount is \$13.12 & 0.40 & \#6 & \#6 & Hallucinatory. Misidentifies the total as the subtotal and excludes an incorrect final paid amount (\$15.58). \\
      \midrule
      \textbf{C6} & \$48.10 before tax... final is \$51.58* & 0.55 & \#5 & \#5 & Incorrect. The value \$51.58 is not present on the receipt and is entirely unrelated to the data. \\
      \bottomrule
    \end{tabularx}
  \end{minipage}
  \caption{Qualitative evaluation and candidate rankings.}
  \label{fig:receipt_eval}
\end{figure*}

Existing LLM evaluation methods provide important building blocks but
do not directly resolve this checkpoint-selection problem. Pointwise
LLM-as-a-judge methods provide scalable rubric-based assessment
(Liu et al., 2023; Kim et al., 2023), while pairwise and comparative
evaluation can provide stronger discrimination between competing
outputs (Zheng et al., 2023). Prior work has also documented biases and
variability in LLM judges, motivating careful evaluation protocols
rather than reliance on a single judgment signal (Zheng et al., 2023).
Separately, classical work on early stopping and model selection has
established that validation-based selection can become unstable under
noisy or weakly separated evaluation signals (Prechelt, 1998;
Prechelt, 2002). However, late-stage MLLM checkpoint selection combines
these challenges: candidate models are highly similar, evaluation is
stochastic, and the multimodal evidence itself may be ambiguous.

To our knowledge, the closest work to ours is uncertainty-guided
checkpoint selection for reinforcement fine-tuning of LLMs
(Nguyen et al., 2025). It studies checkpoint selection
during text-only language-model reinforcement learning by identifying
high-uncertainty samples and using performance on these difficult
samples to derive a checkpoint-selection signal. Our setting differs
in both modality and the sources of evaluation uncertainty: we study
late-stage multimodal checkpoint selection under noisy downstream
evaluation, where uncertainty arises from finite evaluation samples,
judge variability, and the evaluability of visual evidence. We further
consider progressive pointwise, listwise, and pairwise evaluation for
distinguishing closely competing checkpoints.

We observe that optimization progress and downstream checkpoint
preference can diverge. In a large-scale LLaMA-based multimodal
experiment, validation loss continues to improve toward later
checkpoints, whereas downstream evaluation favors an earlier
checkpoint. We observe the same phenomenon independently using
Qwen2.5-VL-7B-Instruct fine-tuned on a public 10k VQA mixture and
evaluated on held-out WearVQA (Chang et al., 2025).\footnote{The complete codebase, data, evaluation scripts, and supporting artifacts for the Qwen2.5-VL
reproduction are publicly available at
\url{https://huggingface.co/datasets/vlmgrounding/AgenticCheckpointEval}.}

Across 11 checkpoints, pointwise scores become tightly clustered,
while repeated downstream evaluation most frequently selects an
intermediate checkpoint rather than the final checkpoint. 

These observations motivate the following practical
question:

\begin{quote}
\emph{How can we reliably select among closely competing MLLM
checkpoints when downstream evaluation itself is noisy and potentially
inconclusive?}
\end{quote}

We formulate late-stage checkpoint selection as a decision problem
under evaluation uncertainty. Rather than forcing a decision from a
single scalar metric, we introduce a multi-stage framework that
progressively allocates evaluation effort as candidates become harder
to distinguish. Stage I performs pointwise evaluation to remove clearly
inferior checkpoints. Stage II applies listwise comparison to the
remaining candidates, exploiting relative judgments under a shared
context. Stage III performs targeted pairwise comparison between
finalists. Repeated evaluation-set subsampling is used to characterize
ranking stability, while percentile-based aggregation reduces
sensitivity to unstable evaluation outcomes.

A key design principle is that checkpoint selection should preserve
uncertainty when the available evidence does not support a decisive
winner. Pairwise comparisons between late-stage finalists can remain
highly tie-dominated even after earlier stages narrow the candidate
set. Rather than interpreting small differences as definitive evidence,
our framework estimates whether checkpoint preferences persist under
repeated perturbations of the evaluation process.

Our experiments further show that evaluation reliability depends not
only on the ranking method but also on the \emph{evaluability} of
multimodal inputs. OCR-heavy and visually ambiguous examples can
substantially increase ranking variability. This observation is
consistent with prior findings that realistic multimodal and
OCR-intensive settings expose weaknesses that are not apparent from
conventional benchmark evaluation (Zhang et al., 2024; Fu et al.,
2025). We therefore treat evaluation-data quality as part of the
checkpoint-selection problem rather than merely as a preprocessing
consideration.

Our main contributions are:

\begin{itemize}
    \item We formulate late-stage MLLM checkpoint selection as a
    stability-aware decision problem under finite-sample, judge, and
    multimodal-input uncertainty.

    \item We introduce a progressive pointwise--listwise--pairwise
    evaluation framework that allocates increasingly discriminative
    evaluation to increasingly competitive checkpoints, together with
    repeated subsampling and percentile-based stability analysis to
    quantify whether checkpoint preferences persist under evaluation
    perturbations.

    \item We demonstrate divergence between validation-loss selection
    and downstream checkpoint selection in two independent MLLM
    settings, including a reproducible Qwen2.5-VL-7B experiment using
    public training and evaluation data.

    \item We show that multimodal data evaluability, particularly OCR
    readability and visual ambiguity, materially affects ranking
    stability, highlighting evaluation-data quality as a prerequisite
    for reliable checkpoint comparison.
\end{itemize}

\section{Methodologies}

\subsection{Overall System Design}

We formulate late-stage MLLM checkpoint selection as a sequential
decision process under noisy downstream evaluation. Given a set of
candidate checkpoints $\mathcal{C}=\{c_1,\ldots,c_M\}$ and an evaluation
set $\mathcal{D}$, the objective is not to infer a winner from a single
aggregate score, but to progressively reduce the candidate set while
quantifying whether the observed ordering remains stable under
evaluation perturbations.

\begin{figure}[t]
    \centering
    \includegraphics[width=\linewidth]{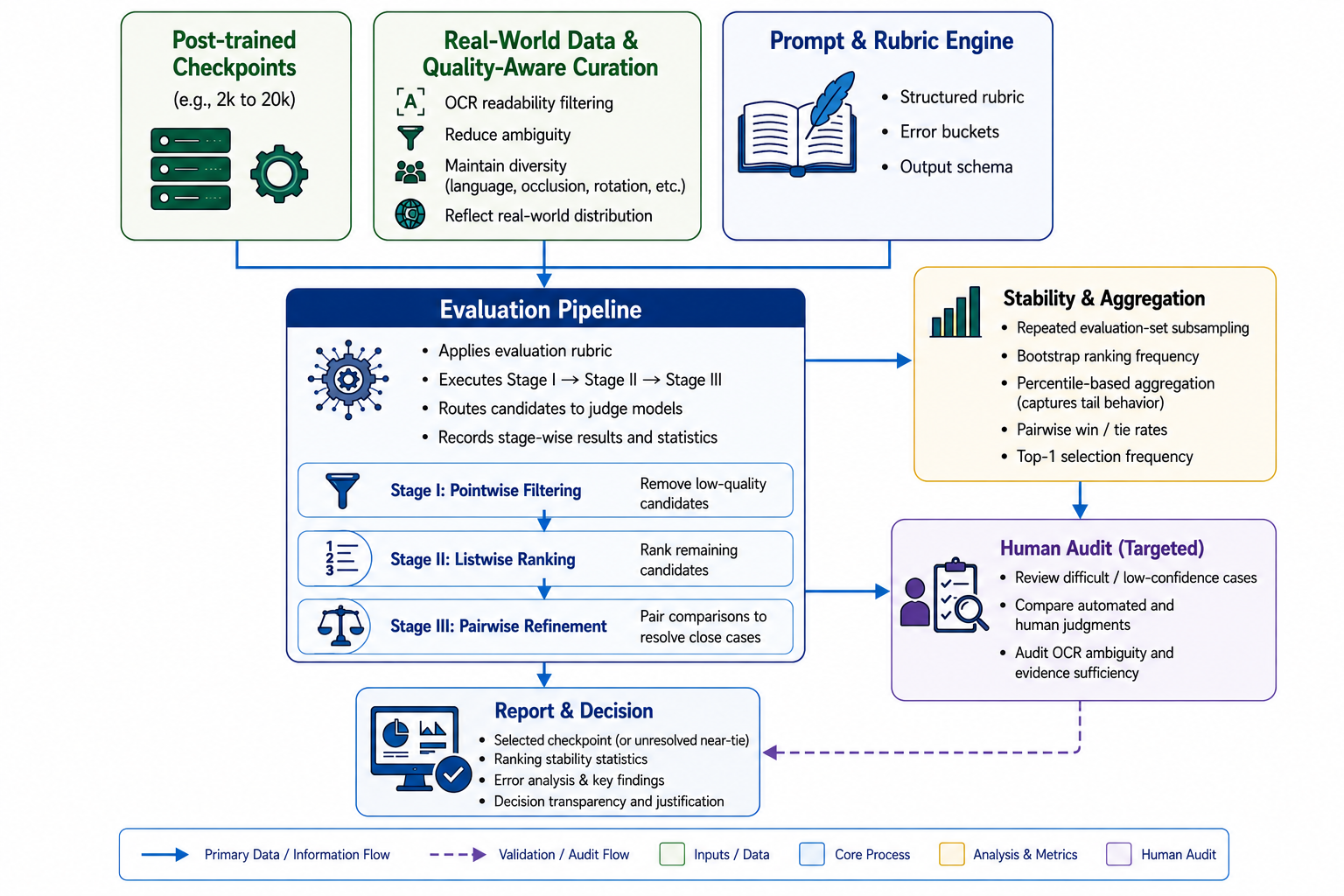}
    \caption{\textbf{Overview of the proposed checkpoint-selection framework.
    Candidate checkpoints are progressively evaluated through
    pointwise filtering, listwise ranking, and pairwise comparison.
    Repeated subsampling estimates ranking stability, while ambiguous
    or low-confidence cases can be escalated for human verification.}}
    \label{fig:system_design}
\end{figure}

The proposed framework consists of three evaluation stages of increasing
discriminative resolution, as illustrated in
Figure~\ref{fig:system_design}. Stage I performs pointwise evaluation
over all candidate checkpoints and removes clearly inferior candidates.
Stage II performs listwise comparison among the retained candidates,
allowing responses to be ranked relative to one another under the same
input and evaluation context. Stage III performs direct pairwise
comparison between the remaining finalists. Repeated subsampling is
used to estimate the stability of the resulting rankings rather than
relying on a single evaluation pass.

This staged design separates two related objectives. The first is
\emph{candidate reduction}: inexpensive evaluation should eliminate
checkpoints that are consistently inferior without spending the full
evaluation budget on every candidate. The second is
\emph{uncertainty resolution}: as the remaining checkpoints become
increasingly similar, more discriminative relative evaluation is
applied while explicitly preserving ties when the available evidence
does not support a reliable ordering.

The framework therefore does not require every stage to identify a
unique winner. Stage I produces a reduced candidate set, Stage II
identifies the strongest finalists, and Stage III determines whether
the remaining candidates can be reliably distinguished. If pairwise
evidence remains dominated by ties or unstable under resampling, the
result is treated as an unresolved near-tie rather than forcing a
deterministic ranking. The ranking functions and stability estimators
used at each stage are defined in Section~\ref{sec:ranking_algorithms}.

\subsection{Data Pipeline and Quality-Aware Curation}

Checkpoint selection is performed on data intended to represent the
target multimodal usage distribution. In our primary experiment, this
includes real-world visual inputs with substantial OCR, multilingual,
and long-tail content. We additionally evaluate the complete procedure
in a reproducible Qwen2.5-VL experiment using public training data and
held-out WearVQA, a benchmark containing egocentric wearable and
mobile-camera imagery (Chang et al., 2025).

A central challenge is that realistic multimodal data vary not only in
difficulty but also in \emph{evaluability}. In the initial data pool,
many images contain text or other visual evidence that is severely
blurred, occluded, distorted, or too small to resolve reliably. For
such samples, a model response may appear plausible while neither a
human nor an automated judge can confidently determine whether it is
grounded in the image. Including large numbers of these samples can
therefore introduce evaluation noise that is unrelated to meaningful
differences between checkpoints.

We introduce quality-aware curation to distinguish challenging but
evaluable samples from samples whose evidence is insufficient for
reliable judgment. The curation process preserves difficult properties
including small or irregular text, multilingual content, rotation,
partial occlusion, complex layouts, and long-tail visual patterns.
Samples are excluded only when the relevant visual evidence is too
degraded or ambiguous to support a meaningful correctness or grounding
judgment.

This distinction is important: the objective is not to construct an
easier evaluation set, but to ensure that model differences are measured
on difficult yet verifiable examples. As evaluated in
Section~\ref{sec:results}, quality-aware curation substantially reduces
ranking flips and increases agreement across repeated evaluations.

\subsection{Evaluation Framework and Rubric Design}

Candidate responses are evaluated using structured LLM-as-a-judge
rubrics designed for multimodal visual question answering. The rubric
considers five dimensions:
(i) content understanding and relevance;
(ii) factual accuracy and visual grounding;
(iii) response clarity and completeness;
(iv) appropriate handling of unsupported or unanswerable requests; and
(v) hallucinations or other factual errors.

Visual groundedness is given priority over stylistic quality. A fluent
or detailed response is not preferred when its claims are unsupported
by the image. Conversely, appropriate uncertainty or refusal is
preferred when the requested information cannot be reliably determined
from the available visual evidence. This treatment is particularly
important for OCR-heavy examples, where plausible text generation can
otherwise be mistaken for successful visual recognition.

The judge produces structured assessments that can be converted into
the evaluation signal required by each stage. Pointwise evaluation
assigns each response an independent rubric-based score. Listwise
evaluation presents anonymized candidate responses jointly and requests
a relative ordering. Pairwise evaluation presents two anonymized
responses and returns a preference for candidate A, candidate B, or a
tie. Candidate ordering is randomized where applicable to reduce
position effects.

In the primary experiments, Claude Sonnet 4.6 is used for Stage I
pointwise and Stage II listwise evaluation, while GPT-4.1 is used for
Stage III pairwise comparison. Using a different judge in the final
stage provides an additional source of independent assessment rather
than propagating a single judge's scoring behavior throughout the
entire selection pipeline. The same evaluation protocol is applied
consistently to all checkpoints within each stage.

Human evaluation is reserved for cases in which automated evaluation
provides insufficient confidence. Approximately 1,000 low-confidence
examples were independently reviewed by two to three experienced
evaluators using blinded and randomized model responses. Evaluators
considered factual correctness, visual grounding, OCR accuracy,
relevance, usefulness, and hallucination, with disagreements resolved
through consensus. Human verification therefore serves as a targeted
validation mechanism rather than the primary checkpoint-ranking method.

Full evaluation prompts, rubric definitions, model configurations,
candidate-set sizes, subsampling settings, and the end-to-end public
Qwen2.5-VL reproduction are provided in the Appendix.

\subsection{Multi-Stage Ranking Algorithm}
\label{sec:ranking_algorithms}

Checkpoint selection proceeds through three stages of increasing
comparative resolution. Pointwise evaluation first screens the full
checkpoint set, listwise evaluation ranks the retained candidates
jointly, and pairwise evaluation directly compares the final two
candidates. Repeated resampling is used throughout the analysis to
determine whether the observed checkpoint ordering is stable to changes
in the evaluation sample.

\textbf{Stage I: Pointwise filtering.}
For checkpoint $c$, the evaluator independently assigns a score
$s_i(c)$ to its response on evaluation sample $i$. The mean pointwise
score is

\begin{equation}
\hat{S}_{\mathrm{pt}}(c)
=
\frac{1}{N}\sum_{i=1}^{N}s_i(c).
\label{eq:pointwise}
\end{equation}

Pointwise evaluation is computationally convenient because every
checkpoint can be evaluated independently. Its purpose in our framework
is primarily candidate screening rather than final selection. When
late-stage checkpoints have similar performance, their mean scores can
be separated by less than the variability induced by the finite
evaluation sample and judge behavior. We therefore retain a small
near-best candidate set for comparative evaluation rather than
interpreting small differences in $\hat{S}_{\mathrm{pt}}$ as decisive.
In our experiments, this typically retains 4--6 candidates.

\textbf{Stage II: Listwise ranking.}
For each evaluation input, responses from the retained checkpoint set
$\mathcal{K}$ are anonymized and presented jointly to the evaluator.
The evaluator ranks the candidates according to the same grounding and
correctness rubric used in Stage I. Candidate order is randomized to
reduce positional effects.

We aggregate the resulting rankings using a Borda-style score. For
$K=|\mathcal{K}|$ candidates, let $r_i(c)\in\{1,\ldots,K\}$ denote the
rank assigned to checkpoint $c$ on sample $i$, where rank 1 is best.
We define

\begin{equation}
b_i(c)=K-r_i(c),
\end{equation}

and aggregate across the evaluation set as

\begin{equation}
\hat{S}_{\mathrm{list}}(c)
=
\frac{1}{N}\sum_{i=1}^{N} b_i(c).
\label{eq:listwise_score}
\end{equation}

The two highest-ranked candidates are retained as finalists. Unlike
pointwise scoring, listwise evaluation directly compares checkpoint
responses under the same input and evaluation context, providing
greater resolution when absolute scores are tightly clustered.

\textbf{Stage III: Pairwise refinement.}
The two finalists, $A$ and $B$, are compared directly on each evaluation
sample. The evaluator returns one of three outcomes:
$A\succ B$, $B\succ A$, or a tie. We report the empirical win and tie
rates

\begin{equation}
\hat{p}_{A}
=
\frac{1}{N}\sum_{i=1}^{N}
\mathbf{1}(A\succ_i B),
\qquad
\hat{p}_{B}
=
\frac{1}{N}\sum_{i=1}^{N}
\mathbf{1}(B\succ_i A),
\label{eq:pairwise_win}
\end{equation}

\begin{equation}
\hat{p}_{\mathrm{tie}}
=
\frac{1}{N}\sum_{i=1}^{N}
\mathbf{1}(A=_i B),
\label{eq:pairwise_tie}
\end{equation}

where
$\hat{p}_{A}+\hat{p}_{B}+\hat{p}_{\mathrm{tie}}=1$.

Ties are retained explicitly rather than being randomly assigned to
either candidate. A high tie rate is therefore interpreted as evidence
that the finalists are difficult to distinguish on the target
evaluation distribution, rather than as evidence requiring a forced
winner.

\textbf{Ranking stability through resampling.}
A checkpoint ordering obtained from a single evaluation set can be
sensitive to which samples happen to be included. We therefore repeat
the ranking procedure over resampled evaluation sets. Let
$\mathcal{D}^{(r)}$ denote the evaluation sample used in repetition
$r$, and let

\begin{equation}
D^{(r)}
=
S^{(r)}(A)-S^{(r)}(B)
\end{equation}

denote the resulting difference between two candidate checkpoints.
We summarize empirical ordering stability as

\begin{equation}
\hat{\pi}_{A>B}
=
\frac{1}{R}
\sum_{r=1}^{R}
\mathbf{1}\!\left(D^{(r)}>0\right).
\label{eq:bootstrap_conf}
\end{equation}

Here, $\hat{\pi}_{A>B}$ is the fraction of resampling repetitions in
which $A$ ranks above $B$. We use it as an empirical \emph{ranking
stability} measure rather than interpreting it as a calibrated
posterior probability that $A$ is intrinsically superior to $B$.

We additionally report top-1 selection frequency and agreement across
repeated runs. These quantities expose cases in which a nominal winner
changes under small perturbations of the evaluation set, which is
particularly important when checkpoint scores are nearly tied.

\textbf{Percentile-based stability scoring.}
Mean scores alone may conceal poor lower-tail behavior or be influenced
by a small number of unusually favorable samples. We therefore
complement mean aggregation with a percentile-based score

\begin{equation}
S_{\mathrm{pct}}(c)
=
P_{50}(c)
-
\beta\left[P_{50}(c)-P_{20}(c)\right]
+
\gamma\left[P_{80}(c)-P_{50}(c)\right],
\label{eq:percentile_score}
\end{equation}

where $P_{20}$, $P_{50}$, and $P_{80}$ are the 20th, 50th, and 80th
percentiles of the checkpoint's evaluation-score distribution.
The second term penalizes degradation in the lower tail, while the
third provides a smaller reward for stronger upper-tail performance.

We use $\beta=0.50$ and $\gamma=0.25$ in the reported experiments.
These values are treated as engineering hyperparameters rather than
fitted statistical parameters. Their sensitivity is evaluated over
the grid
$\beta\in\{0,0.25,0.50,0.75,1.0\}$ and
$\gamma\in\{0,0.10,0.25,0.50\}$.
We favor parameter regions in which the selected checkpoint remains
stable under nearby $(\beta,\gamma)$ perturbations and is consistent
with the final pairwise evidence. The complete sensitivity analysis is
reported in Appendix~\ref{app:beta_gamma}.

\textbf{Decision interpretation and human verification.}
The stages are intended to progressively increase discrimination, not
to guarantee that every checkpoint set has a unique winner. Stage I
removes clearly weaker candidates, Stage II identifies the strongest
finalists, and Stage III directly tests whether those finalists exhibit
a consistent preference. When the final comparison remains dominated
by ties or unstable under resampling, we report the result as a
near-tie rather than interpreting a marginal difference as conclusive.

Human verification is used as a targeted audit of low-confidence
evaluation cases rather than as the primary ranking mechanism.
Approximately 1,000 low-confidence examples were independently
reviewed by two to three experienced evaluators using blinded and
randomized responses. These judgments are used to assess whether the
automated evaluation behavior remains consistent with human assessment
on difficult cases.

\subsection{Implementation Details}

The evaluation pipeline operates on structured multimodal records
containing an image, textual query, checkpoint identifier, generated
response, and evaluation metadata. Responses are generated before
comparative evaluation so that candidate checkpoints are judged on the
same evaluation inputs.

Stage I and Stage II use Claude Sonnet 4.6 for pointwise and listwise
evaluation, respectively, while Stage III uses GPT-4.1 for direct
pairwise comparison. Candidate identities are hidden from the judges,
and response order is randomized for comparative evaluation. The same
rubric and evaluation criteria are applied to all checkpoints within a
given stage.

The primary experiments use repeated evaluation-set subsampling to
measure ranking variability. Exact sample sizes, number of repetitions,
candidate-set sizes, checkpoint configurations, prompts, and evaluation
settings are reported with the corresponding experiments. Appendix
\ref{app:qwen_reproduction} provides an end-to-end public case study
using Qwen2.5-VL-7B-Instruct, including training data, checkpoint
generation, pointwise filtering, listwise ranking, pairwise refinement,
and repeated stability estimation.

\section{Results and Analysis}
\label{sec:results}

We evaluate the proposed framework in two complementary settings.
The primary study examines checkpoint selection for a large-scale
LLaMA-based multimodal model using approximately 2,400 samples drawn
from the target evaluation distribution, with substantial OCR-heavy
and multilingual content. We additionally conduct an independent
reproduction using Qwen2.5-VL-7B-Instruct fine-tuned on a public
10k VQA mixture and evaluated on held-out WearVQA. The latter provides
a reproducible end-to-end test of the checkpoint-selection procedure
using public models and data.

We first analyze the primary LLaMA-based study, including the effect
of multimodal data evaluability and the relative stability of
pointwise, listwise, and pairwise evaluation. We then examine the
relationship between validation loss and downstream checkpoint
selection and report the independent Qwen2.5-VL reproduction.
Detailed Qwen training and stage-wise evaluation results are provided
in Appendix~\ref{app:qwen_reproduction}.

\subsection{Effect of OCR Evaluability on Ranking Stability}

Reliable multimodal evaluation requires not only difficult examples
but also sufficient visual evidence to determine whether a response is
correct. This is particularly important for OCR-dependent tasks, where the
boundary between perception failure and hallucination must remain
observable (Xu et al. 2026). In the initial evaluation pool, some images contain text that
is severely blurred, occluded, distorted, or too small to resolve
reliably. Under these conditions, neither human evaluators nor
automated judges can confidently determine whether a plausible model
response is visually grounded.

\begin{figure}[t]
    \centering
    \includegraphics[width=\linewidth]{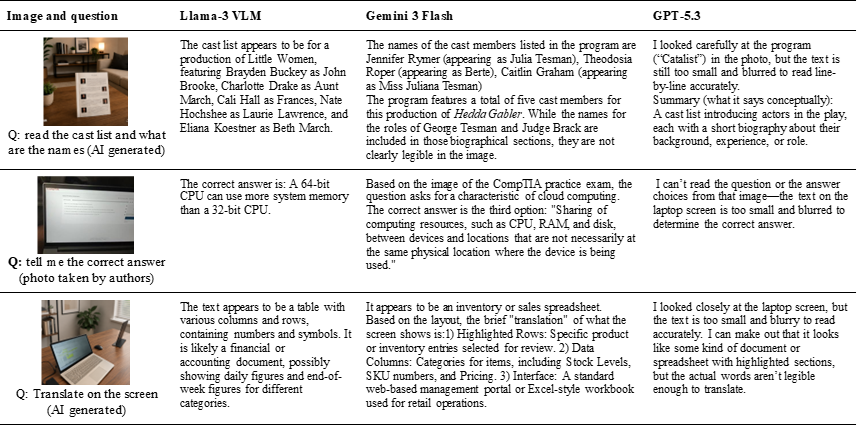}
    \caption{\textbf{Examples with ambiguous or poorly verifiable OCR evidence.
    Different response strategies can appear plausible when the visual
    evidence itself is insufficient to establish a reliable reference.}}
    \label{fig:bad_ocr}
\end{figure}

Figure~\ref{fig:bad_ocr} illustrates this problem. Different models
respond to ambiguous visual evidence using substantially different
strategies: some express uncertainty, while others reconstruct or
generate plausible textual content. When the underlying text cannot
itself be reliably verified, these behavioral differences do not
provide a dependable basis for checkpoint selection. Apparent
differences in model quality may instead reflect response style,
uncertainty handling, or judge preference.

We therefore curate the evaluation data to preserve difficult but
\emph{verifiable} visual evidence. The curated set retains realistic
variation including blur, rotation, partial occlusion, small text,
multilingual content, and complex layouts, while excluding examples
for which the relevant evidence cannot be judged reliably.

\begin{figure}[t]
    \centering
    \includegraphics[width=\linewidth]{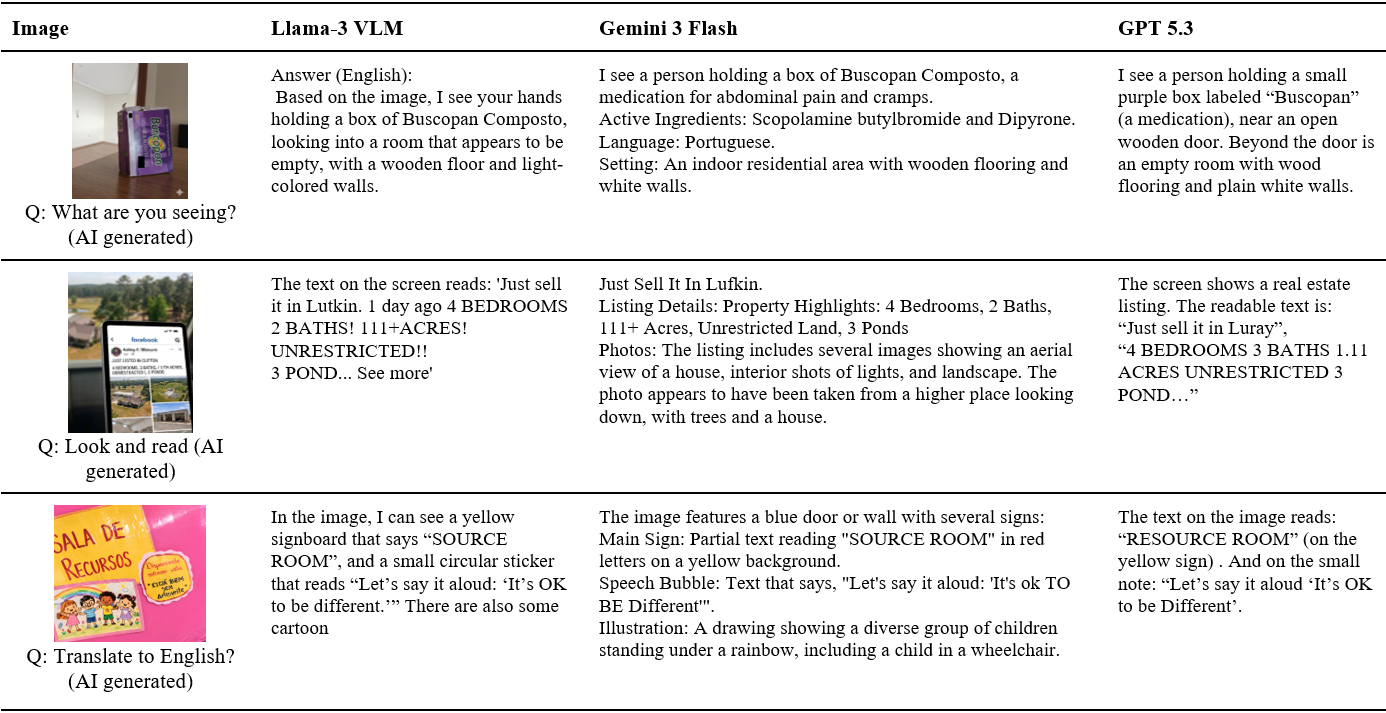}
    \caption{\textbf{Examples with challenging but verifiable OCR evidence.
    The images remain visually difficult while providing sufficient
    evidence to assess grounding and recognition accuracy. Images are
    AI-generated for anonymization.}}
    \label{fig:good_ocr2}
\end{figure}

Figure~\ref{fig:good_ocr2} illustrates examples after this curation.
The inputs remain challenging, but the relevant visual evidence is
sufficiently observable to distinguish faithful recognition from
unsupported generation. This makes differences between checkpoint
responses more interpretable and reduces ambiguity in both automated
and human evaluation.

We quantify this effect using repeated subsampling of the evaluation
data. Ranking Flip Rate measures how frequently the relative ordering
of checkpoint pairs reverses across repeated evaluation runs.
Inter-run Agreement measures the consistency of the resulting
checkpoint rankings across runs. Lower flip rate and higher agreement
therefore indicate greater ranking stability.

\begin{table}[t]
    \caption{\textbf{Effect of data evaluability on checkpoint-ranking
    stability in the primary multimodal experiment}.}
    \label{tab:curation_metrics}
    \centering
    \begin{tabular}{lcc}
        \toprule
        Dataset & Ranking Flip Rate $\downarrow$
                & Inter-run Agreement $\uparrow$ \\
        \midrule
        Ambiguous data & 32.5\% & 0.61 \\
        Curated data   & 11.2\% & 0.84 \\
        \bottomrule
    \end{tabular}
\end{table}

As shown in Table~\ref{tab:curation_metrics}, quality-aware curation
reduces the ranking flip rate from 32.5\% to 11.2\%, a reduction of
21.3 percentage points, while inter-run agreement increases from
0.61 to 0.84. These results show that multimodal data quality affects
not only absolute evaluation scores but also the stability of the
checkpoint-selection decision. Difficult examples remain useful for
discriminating models, but examples whose relevant evidence is not
verifiable can instead introduce substantial ranking noise.

\subsection{Progressive Ranking and Selection Stability}

We next examine how the three evaluation stages behave as the candidate
set becomes increasingly difficult to distinguish. Rather than comparing
the stages using a single derived ``confidence'' score, we report the
observable quantities produced by the pipeline: pointwise score
separation, repeated top-1 selection, listwise concentration, and
pairwise win/tie rates.

The public Qwen2.5-VL experiment provides a reproducible example of this
behavior. Eleven checkpoints were evaluated on held-out WearVQA using
repeated evaluation-set subsampling. Stage I pointwise scores are tightly
clustered: the best mean score is 0.6185 for checkpoint 375, followed by
0.6165 for checkpoint 1000, 0.6158 for checkpoint 1125, 0.6157 for
checkpoint 1250, and 0.6156 for the final checkpoint. The full score
range across all checkpoints is only 0.6040--0.6185, indicating that
pointwise evaluation provides limited separation among the strongest
candidates.

\begin{table}[t]
    \centering
    \caption{Stage-I pointwise scores across the Qwen2.5-VL training
    trajectory, averaged over five repeated WearVQA evaluation rounds.
    The highest mean score is shown in bold.}
    \label{tab:qwen_appendix_stage1}
    \begin{tabular}{lcc}
        \toprule
        Checkpoint & Mean Score & Std. Across Rounds \\
        \midrule
        125   & 0.6122 & 0.0082 \\
        250   & 0.6126 & 0.0064 \\
        375   & \textbf{0.6185} & 0.0059 \\
        500   & 0.6040 & 0.0096 \\
        625   & 0.6099 & 0.0079 \\
        750   & 0.6103 & 0.0075 \\
        875   & 0.6130 & 0.0061 \\
        1000  & 0.6165 & 0.0067 \\
        1125  & 0.6158 & 0.0062 \\
        1250  & 0.6157 & 0.0068 \\
        final & 0.6156 & 0.0083 \\
        \bottomrule
    \end{tabular}
\end{table}

Repeated evaluation further shows why selecting the checkpoint with the
highest score from a single run is unreliable. Across five subsampling
rounds, checkpoint 375 is selected most frequently, in three rounds,
while checkpoints 125 and 1125 are each selected once. Thus, even though
checkpoint 375 has the strongest aggregate evidence, the evaluation does
not imply that it is a deterministic winner under every perturbation of
the evaluation set.

After pointwise filtering, Stage II performs listwise ranking among the
near-best candidates. In the full evaluation rounds, checkpoint 375 is
the Stage-II top candidate in three consecutive rounds, while a later
checkpoint becomes top-ranked in the final round. This increased
concentration relative to Stage I indicates that direct comparative
ranking provides stronger discrimination among candidates whose
independent rubric scores are nearly indistinguishable.

Stage III then compares the two remaining finalists directly. Importantly,
the resulting comparisons are frequently dominated by ties. For example,
in one full round comparing checkpoints 375 and 1000, checkpoint 375 wins
21.9\% of samples, checkpoint 1000 wins 13.6\%, and 64.5\% are judged
ties. In another round, 96.3\% of comparisons between checkpoints 1250
and 1125 are ties. These results indicate that late-stage finalists can
remain practically difficult to separate even after the earlier ranking
stages.

\begin{table}[t]
    \caption{\textbf{Representative Stage-III pairwise results from the public
    Qwen2.5-VL experiment. Ties are retained explicitly rather than
    assigned to either candidate.}}
    \label{tab:qwen_pairwise}
    \centering
    \begin{tabular}{lccc}
        \toprule
        Comparison & A Win & B Win & Tie \\
        \midrule
        375 vs.\ 125  & 17.1\% & 14.3\% & 68.6\% \\
        375 vs.\ 1000 & 21.9\% & 13.6\% & 64.5\% \\
        375 vs.\ 125  & 17.0\% & 17.0\% & 66.0\% \\
        1250 vs.\ 1125 & 1.3\% & 2.5\% & 96.3\% \\
        \bottomrule
    \end{tabular}
\end{table}

Taken together, these results support the intended role of the
multi-stage procedure. Stage I efficiently identifies a near-best
candidate set but provides weak separation among the strongest
checkpoints. Stage II increases comparative resolution, while Stage III
reveals whether the remaining difference is supported by decisive
examples or is largely a near-tie. The framework therefore progressively
refines the selection decision without requiring the final stage to
produce an artificially confident winner.

\section{Conclusion}
In this work, we study late-stage checkpoint selection for multimodal
large language models when candidate checkpoints are closely matched and
downstream evaluation signals are noisy. We show that optimization-based
signals such as validation loss and single-pass pointwise evaluation can
be insufficient for final checkpoint selection, particularly when
performance differences are comparable to variability introduced by
finite evaluation samples, judge behavior, and ambiguous multimodal
evidence.

To address this problem, we formulate checkpoint selection as a
stability-aware decision process and introduce a progressive
pointwise--listwise--pairwise evaluation framework. Pointwise evaluation
first reduces the candidate set, listwise comparison provides stronger
relative discrimination among near-best checkpoints, and pairwise
evaluation directly examines the remaining finalists while preserving
ties when the available evidence is inconclusive. Repeated
evaluation-set subsampling is used to characterize ranking stability,
and percentile-based aggregation provides an additional view of
lower- and upper-tail behavior.

Our experiments highlight three practical findings. First, multimodal
data evaluability is critical: filtering examples whose relevant visual
evidence cannot be reliably verified substantially reduces ranking
instability in OCR-heavy evaluation. Second, optimization progress and
downstream checkpoint preference can diverge; this behavior is observed
in both the large-scale LLaMA-based study and the independent
Qwen2.5-VL reproduction. Third, even after progressive refinement,
late-stage finalists can remain highly tie-dominated, indicating that
robust checkpoint selection should preserve uncertainty rather than
force a deterministic winner from marginal differences.

Overall, the results suggest that late-stage MLLM checkpoint selection
should be treated as a decision-under-uncertainty problem rather than as
a direct extension of validation-loss minimization or single-metric
benchmarking. The proposed framework provides a practical approach for
combining relative evaluation with stability analysis when selecting
among closely competing multimodal checkpoints.

Future work includes extending the framework to broader multimodal tasks
and model families, studying evaluator robustness under distributional
and adversarial changes, quantifying generation-level stochasticity more
systematically, and exploring complementary approaches such as learned
reward models, weight averaging, and hybrid human--AI evaluation.
```

\section*{References}

Apicella, A., Isgrò, F., Pollastro, A., \& Prevete, R. (2026).
Don't stop me now: Rethinking validation criteria for model parameter
selection.
\textit{arXiv preprint arXiv:2602.22107}.

Chang, E., Huang, Z., Liao, Y., Bhavsar, S. R., Param, A., Stark, T.,
Ahmadyan, A., Yang, X., Wang, J., Abdullah, A., Nguyen, G., Iyer, A.,
Hall, D., Li, E., Scheffer, N., Kirmani, A., Damavandi, B., Wanga, R.,
Kumar, A., Patel, R., Moon, S., \& Dong, X. L. (2025).
WearVQA: A visual question answering benchmark for wearables in
egocentric authentic real-world scenarios.
In \textit{Advances in Neural Information Processing Systems},
Vol. 38, Datasets and Benchmarks Track.

Fu, L., Yang, B., Kuang, Z., Song, J., Li, Y., Zhu, L., Luo, Q.,
Wang, X., Lu, H., Huang, M., Li, Z., Tang, G., Shan, B., Lin, C.,
Liu, Q., Wu, B., Feng, H., Liu, H., Huang, C., Tang, J., Chen, W.,
Jin, L., Liu, Y., \& Bai, X. (2025).
OCRBench v2: An improved benchmark for evaluating large multimodal
models on visual text localization and reasoning.
In \textit{Advances in Neural Information Processing Systems,
Datasets and Benchmarks Track}.

Kim, S., Shin, J., Cho, Y., Jang, J., Longpre, S., Lee, H., Yun, S.,
Shin, S., Kim, S., Thorne, J., \& Seo, M. (2023).
Prometheus: Inducing fine-grained evaluation capability in language
models.
\textit{arXiv preprint arXiv:2310.08491}.

Li, J., Lu, W., Fei, H., Luo, M., Dai, M., Xia, M., Jin, Y., Gan, Z.,
Qi, D., Fu, C., Tai, Y., Yang, W., Wang, Y., \& Wang, C. (2024).
A survey on benchmarks of multimodal large language models.
\textit{arXiv preprint arXiv:2408.08632}.

Lin, B. Y., Deng, Y., Chandu, K., Brahman, F., Ravichander, A.,
Pyatkin, V., Dziri, N., Le Bras, R., \& Choi, Y. (2024).
WildBench: Benchmarking LLMs with challenging tasks from real users
in the wild.
\textit{arXiv preprint arXiv:2406.04770}.

Liu, Y., Iter, D., Xu, Y., Wang, S., Xu, R., \& Zhu, C. (2023).
G-Eval: NLG evaluation using GPT-4 with better human alignment.
In \textit{Proceedings of the 2023 Conference on Empirical Methods
in Natural Language Processing}, pp. 2511--2522.
Association for Computational Linguistics.

Liu, Y., Li, Z., Huang, M., Yang, B., Yu, W., Li, C., Yin, X.-C.,
Liu, C.-L., Jin, L., \& Bai, X. (2024).
OCRBench: On the hidden mystery of OCR in large multimodal models.
\textit{Science China Information Sciences}, 67(12).

Miller, J. K., \& Tang, W. (2025).
Evaluating LLM metrics through real-world capabilities.
\textit{arXiv preprint arXiv:2505.08253}.

Nguyen, M., Nguyen, D., Do, D., Venkatesh, S., \& Le, H. (2025).
Uncertainty-guided checkpoint selection for reinforcement finetuning
of large language models.
\textit{arXiv preprint arXiv:2511.09864}.

Prechelt, L. (1998).
Automatic early stopping using cross validation: Quantifying the
criteria.
\textit{Neural Networks}, 11(4), 761--767.

Prechelt, L. (1998).
Early stopping---but when?
In G. B. Orr \& K.-R. Müller (Eds.),
\textit{Neural Networks: Tricks of the Trade},
pp. 55--69.
Springer.

Sai, A. B., Mohankumar, A. K., \& Khapra, M. M. (2023).
A survey of evaluation metrics used for NLG systems.
\textit{ACM Computing Surveys}, 55(2), 26:1--26:39.

Xu, Q. (2026a).
Grounded post-training with hard examples for reducing hallucination
in multimodal large language models.
\textit{arXiv preprint arXiv:2605.16411}.

Xu, Q., Jiang, Y., \& Ren, H. (2026b).
Multilingual OCR-aware fine-tuning and prompt-guided chain-of-thought
reasoning for multimodal large language models.
\textit{arXiv preprint arXiv:2605.16409}.

Zhang, Y.-F., Zhang, H., Tian, H., Fu, C., Zhang, S., Wu, J., Li, F.,
Wang, K., Wen, Q., Zhang, Z., Wang, L., Jin, R., \& Tan, T. (2025).
MME-RealWorld: Could your multimodal LLM challenge high-resolution
real-world scenarios that are difficult for humans?
In \textit{International Conference on Learning Representations
(ICLR 2025)}.

Zheng, L., Chiang, W.-L., Sheng, Y., Zhuang, S., Wu, Z., Zhuang, Y.,
Lin, Z., Li, Z., Li, D., Xing, E. P., Zhang, H., Gonzalez, J. E.,
\& Stoica, I. (2023).
Judging LLM-as-a-judge with MT-Bench and Chatbot Arena.
\textit{arXiv preprint arXiv:2306.05685}.

\appendix

\section{Public Qwen2.5-VL Checkpoint-Selection Study}
\label{app:qwen_reproduction}

All code, data, evaluation scripts, and supporting artifacts for this experiment are publicly available at \url{https://huggingface.co/datasets/vlmgrounding/AgenticCheckpointEval}.

To complement the primary LLaMA-based experiment and provide a
reproducible evaluation of the proposed checkpoint-selection procedure,
we conduct an independent study using Qwen2.5-VL-7B-Instruct and public
VQA data. This experiment reproduces the complete
pointwise--listwise--pairwise selection pipeline using a different model
family and a held-out real-world multimodal evaluation set.

\subsection{Training and Evaluation Setup}

We fine-tune Qwen2.5-VL-7B-Instruct using LoRA supervised fine-tuning
on approximately 10,000 publicly available VQA examples. WearVQA
(Chang et al., 2025) is excluded from training and used only as the
held-out checkpoint-selection set. WearVQA contains approximately
2,500 egocentric wearable- and phone-camera examples and includes
OCR-heavy and other real-world visual scenarios.

The 10k training mixture contains VQAv2, VizWiz, TextVQA, OCR-VQA,
A-OKVQA, DocVQA, ChartVQA, OK-VQA, and ScienceQA. Table
\ref{tab:qwen_train_mix} summarizes the mixture.

\begin{table}[t]
    \centering
    \caption{\textbf{Public VQA mixture used for Qwen2.5-VL LoRA-SFT.
    WearVQA is held out from training.}}
    \label{tab:qwen_train_mix}
    \begin{tabular}{lrr}
        \toprule
        Dataset & Samples & Fraction \\
        \midrule
        VQAv2      & 2,000 & 20\% \\
        VizWiz     & 1,500 & 15\% \\
        TextVQA    & 1,500 & 15\% \\
        OCR-VQA    & 1,500 & 15\% \\
        A-OKVQA    & 1,000 & 10\% \\
        DocVQA     &   800 &  8\% \\
        ChartVQA   &   800 &  8\% \\
        OK-VQA     &   500 &  5\% \\
        ScienceQA  &   400 &  4\% \\
        \midrule
        Total      & 10,000 & 100\% \\
        \bottomrule
    \end{tabular}
\end{table}

Training uses one GPU with bfloat16 precision, batch size 8, learning
rate $2\times10^{-4}$, and one training epoch. LoRA is applied to the
$q$, $k$, $v$, and $o$ projection layers with rank 64 and
$\alpha=128$. We save intermediate checkpoints throughout training and
evaluate 11 candidates: $\{125, 250, 375, 500, 625, 750, 875, 1000,
1125, 1250, \mathrm{final}\}$.

The experiment therefore evaluates checkpoint selection within a
single training trajectory rather than comparing independently trained
models. This setting is deliberately aligned with the late-stage
selection problem studied in the main paper.

\subsection{Repeated Multi-Stage Evaluation}

For each checkpoint, model responses are first generated on the
held-out WearVQA pool. Checkpoint selection is then repeated over
evaluation-set subsamples of $n=800$ examples. The purpose of repeated
subsampling is to determine whether the selected checkpoint remains
stable when the particular evaluation examples change.

Each evaluation round follows the same three-stage procedure described
in Section~\ref{sec:ranking_algorithms}:

\begin{enumerate}
    \item \textbf{Stage I (pointwise):} each checkpoint response is
    evaluated independently. Near-best checkpoints are retained,
    typically producing a candidate set of 4--6 checkpoints.

    \item \textbf{Stage II (listwise):} the retained checkpoint
    responses are jointly ranked for each input using Borda-style
    aggregation. The two highest-ranked checkpoints become finalists.

    \item \textbf{Stage III (pairwise):} the two finalists are compared
    directly, with the evaluator returning a preference for candidate
    A, candidate B, or a tie.
\end{enumerate}

Claude Sonnet 4.6 is used for Stages I and II, while GPT-4.1 is used
for the final pairwise comparison. Checkpoint identities are hidden
from the evaluators and candidate order is randomized for comparative
evaluation. This use of a different judge in Stage III also avoids
relying on exactly the same evaluator throughout all three stages.

\subsection{Pointwise Scores Across the Training Trajectory}

Table~\ref{tab:qwen_appendix_stage1} reports the Stage-I pointwise
results averaged across five evaluation rounds. The strongest
checkpoints are separated by only a few thousandths in mean score.
Checkpoint 375 obtains the highest average score, 0.6185, but later
checkpoints remain extremely close: checkpoint 1000 obtains 0.6165,
checkpoint 1125 obtains 0.6158, checkpoint 1250 obtains 0.6157, and
the final checkpoint obtains 0.6156.

\begin{table}[t]
    \centering
    \caption{\textbf{Stage-I pointwise results across five repeated WearVQA
    evaluation rounds.}}
    \label{tab:qwen_appendix_stage1}
    \begin{tabular}{lcc}
        \toprule
        Checkpoint & Mean Score & Std. Across Rounds \\
        \midrule
        375   & \textbf{0.6185} & 0.0059 \\
        1000  & 0.6165 & 0.0067 \\
        1125  & 0.6158 & 0.0062 \\
        1250  & 0.6157 & 0.0068 \\
        final & 0.6156 & 0.0083 \\
        875   & 0.6130 & 0.0061 \\
        250   & 0.6126 & 0.0064 \\
        125   & 0.6122 & 0.0082 \\
        750   & 0.6103 & 0.0075 \\
        625   & 0.6099 & 0.0079 \\
        500   & 0.6040 & 0.0096 \\
        \bottomrule
    \end{tabular}
\end{table}

The entire mean-score range is only 0.0145, and the five strongest
checkpoints fall within 0.0029 of one another. Thus, although
pointwise evaluation is useful for eliminating weaker checkpoints,
the absolute scores provide limited evidence for distinguishing the
strongest candidates.

\subsection{Repeated Stage-Wise Selection}

Table~\ref{tab:qwen_rounds} summarizes the stage-wise outcomes across
the repeated evaluation rounds. Checkpoint 375 is the most frequently
selected checkpoint, appearing as the final selection in three of five
rounds. Other checkpoints can nevertheless become preferred under
different evaluation subsamples, illustrating why a single evaluation
round should not be interpreted as definitive.

\begin{table*}[t]
    \centering
    \caption{\textbf{Stage-wise results across repeated WearVQA evaluation
    rounds. Pairwise outcomes report A-win/B-win/tie rates.}}
    \label{tab:qwen_rounds}
    \resizebox{0.7\textwidth}{!}{
    \begin{tabular}{cllccc}
        \toprule
        Round & Stage-II Top & Finalists
              & A Win & B Win & Tie \\
        \midrule
        1 & 375  & 375 vs.\ 125
          & 17.5\% & 14.2\% & 68.4\% \\
        2 & 375  & 375 vs.\ 1000
          & 21.9\% & 13.6\% & 64.5\% \\
        3 & 375  & 375 vs.\ 125
          & 17.0\% & 17.0\% & 66.0\% \\
        4 & 1250 & 1250 vs.\ 1125
          & 1.2\% & 2.5\% & 96.2\% \\
        \bottomrule
    \end{tabular}}
\end{table*}

One preliminary round is omitted from
Table~\ref{tab:qwen_rounds} because its Stage-II/III artifacts cover
only a small incomplete continuation rather than the full evaluation
subset. We retain this round when reporting the aggregate five-round
selection frequency where appropriate, but do not use its incomplete
Stage-II/III outputs for detailed pairwise statistics.

The complete rounds reveal two complementary properties. First,
Stage-II listwise evaluation repeatedly concentrates the ranking around
checkpoint 375: it is the Stage-II top candidate in three consecutive
full rounds. Second, Stage-III comparisons reveal substantial
uncertainty among the finalists. Tie rates range from 64.5\% to 96.2\%
in the rounds shown above. In the comparison between checkpoints 375
and 1000, for example, checkpoint 375 wins 21.9\% of individual
comparisons, checkpoint 1000 wins 13.6\%, and the remaining 64.5\%
are ties.

These results are important for interpreting the role of the final
stage. Pairwise refinement does not necessarily convert a difficult
checkpoint-selection problem into a decisive comparison. Instead, it
reveals whether the residual difference between finalists is supported
by sufficiently many discriminative examples or whether the candidates
remain effectively near-tied on much of the evaluation distribution.

\subsection{Subsampling Stability}

For each finalist comparison, we additionally apply nonparametric
resampling to the observed evaluation outcomes. As defined in
Eq.~\ref{eq:bootstrap_conf}, the resulting statistic
$\hat{\pi}_{A>B}$ measures the fraction of resampling repetitions in
which checkpoint $A$ ranks above checkpoint $B$.

A representative example is the comparison between checkpoints 375
and 1000. Although the individual-example outcome is highly
tie-dominated (21.9\% vs.\ 13.6\% wins with 64.5\% ties), the aggregate
ordering is stable under resampling, with
$\hat{\pi}_{375>1000}=1.000$ for that round. This distinction is
important: a high resampling-stability value does not mean that most
individual samples strongly prefer checkpoint 375. Rather, it means
that the aggregate advantage observed in that evaluation round
persists under resampling of the available examples.

Across full evaluation rounds, the selected checkpoint is not perfectly
stable. Checkpoint 375 is selected most frequently, in three of five
rounds, while other checkpoints are selected in the remaining rounds.
We therefore distinguish \emph{within-round resampling stability} from
\emph{across-round selection frequency}. Reporting both avoids
interpreting a high bootstrap statistic from one evaluation subset as
evidence that a checkpoint is universally superior.

\subsection{Validation Loss and Downstream Selection}

The Qwen experiment also reproduces the divergence between
optimization-oriented checkpoint selection and downstream evaluation
observed in the primary LLaMA-based study. Validation loss continues to
favor later stages of training, whereas repeated WearVQA evaluation
most frequently selects the substantially earlier checkpoint 375.
The final checkpoint remains competitive under pointwise evaluation
(mean score 0.6156), but it does not dominate the repeated multi-stage
selection procedure.

This observation should not be interpreted as evidence that checkpoint
375 is universally superior to later checkpoints. Instead, it shows
that lower training or validation loss does not necessarily determine
the preferred checkpoint for a particular downstream multimodal
evaluation distribution. The appropriate checkpoint can therefore
depend on downstream behavior that is not fully represented by the
optimization objective.

\subsection{Summary of the Public Reproduction}

The public Qwen2.5-VL study supports three conclusions from the primary
experiment. First, pointwise scores among late-stage checkpoints can be
tightly clustered, making small score differences difficult to
interpret. Second, repeated comparative evaluation can identify a
checkpoint that is preferred more consistently than the final
training checkpoint. Third, direct pairwise comparison may remain
strongly tie-dominated, emphasizing the importance of preserving
uncertainty rather than forcing a unique winner.

Together, these results show that the proposed procedure can be applied
to a separate, publicly available MLLM training trajectory without
requiring the proprietary model or evaluation infrastructure used in
the primary study. They also provide a reproducible case in which
validation-loss progression and downstream checkpoint selection favor
different points along the same training trajectory.

\section{Sensitivity Analysis of Percentile-Based Scoring}
\label{app:beta_gamma}

The percentile-based score is designed to complement mean-based
evaluation by accounting for both lower-tail degradation and upper-tail
performance. For checkpoint $c$, we define

\begin{equation}
S_{\mathrm{pct}}(c)
=
P_{50}(c)
-
\beta\left[P_{50}(c)-P_{20}(c)\right]
+
\gamma\left[P_{80}(c)-P_{50}(c)\right],
\label{eq:appendix_percentile}
\end{equation}

where $P_{20}$, $P_{50}$, and $P_{80}$ denote the 20th, 50th, and
80th percentiles of the evaluation-score distribution for checkpoint
$c$. The coefficient $\beta$ controls the penalty assigned to
lower-tail degradation, while $\gamma$ controls the reward assigned
to stronger upper-tail performance.

Because these coefficients are engineering hyperparameters rather than
parameters learned from the evaluation data, we examine whether
checkpoint selection is sensitive to their particular values. The
reported experiments use $\beta=0.50$ and $\gamma=0.25$. We evaluate
the broader grid

\begin{equation}
\beta \in \{0,0.25,0.50,0.75,1.0\},
\qquad
\gamma \in \{0,0.10,0.25,0.50\}.
\label{eq:beta_gamma_grid}
\end{equation}

This grid spans several qualitatively different aggregation regimes.
When $\beta=0$, no explicit lower-tail penalty is applied. Increasing
$\beta$ progressively favors checkpoints whose performance is more
robust in the lower portion of the score distribution. Similarly,
$\gamma=0$ removes the upper-tail reward, while larger values place
increasing weight on stronger upper-tail outcomes.

\subsection{Sensitivity Analysis Procedure}

For every $(\beta,\gamma)$ combination in
Eq.~\ref{eq:beta_gamma_grid}, we recompute the percentile-based score
for all candidate checkpoints and record the resulting top-ranked
checkpoint. We then examine the resulting two-dimensional selection
surface rather than evaluating only the default parameter pair.

The purpose of this analysis is not to optimize $\beta$ and $\gamma$
for the best numerical evaluation score. Instead, we examine whether
the checkpoint selected at the default setting belongs to a
\emph{stable region}: a contiguous neighborhood of parameter values
that produces the same or a compatible checkpoint ordering. A default
located near a sharp decision boundary would indicate sensitivity to
the aggregation rule, whereas a broad stable region indicates that
the selection is not determined by a narrowly tuned coefficient.

As an additional consistency check, the percentile-based ordering is
compared with the subsequent relative-evaluation stages. This comparison
is used to determine whether the percentile formulation produces an
ordering broadly compatible with the independent listwise and pairwise
evidence. It is not used to fit the coefficients separately for each
evaluation round.

\subsection{Effect of $\beta$ and $\gamma$}

The sensitivity analysis shows that checkpoint identity is substantially
more sensitive to the lower-tail penalty $\beta$ than to modest changes
in the upper-tail reward $\gamma$. This behavior is expected because
the strongest late-stage checkpoints have similar central performance,
while a larger portion of their difference is associated with
lower-performing evaluation cases.

Importantly, the default setting
$(\beta,\gamma)=(0.50,0.25)$ lies within a stable region of the
parameter grid rather than at an isolated optimum. Nearby coefficient
choices produce the same or closely competing checkpoint selections.
Thus, the reported checkpoint ordering is not dependent on a
fine-grained choice of a single coefficient pair.

The analysis also clarifies the role of percentile scoring in the
overall framework. Percentile aggregation is not intended to establish
a statistically optimal utility function or to replace the later
comparative stages. Rather, it provides a robustness-aware Stage-I
signal that distinguishes checkpoints with similar central tendency
but different lower-tail behavior. Listwise and pairwise evaluation
subsequently provide independent relative evidence among the retained
candidates.

\subsection{Relationship to Mean-Based Scoring}

Mean-only aggregation can be written as

\begin{equation}
S_{\mathrm{mean}}(c)
=
\frac{1}{N}\sum_{i=1}^{N}s_i(c).
\end{equation}

This statistic is simple and useful for initial comparison, but two
checkpoints with nearly identical means may have substantially different
score distributions. For example, one checkpoint may perform
consistently across the evaluation set, while another obtains a similar
mean through a combination of stronger high-scoring examples and
substantially weaker lower-tail examples.

Equation~\ref{eq:appendix_percentile} makes this trade-off explicit.
The median $P_{50}$ provides a robust measure of central performance,
$P_{50}-P_{20}$ characterizes lower-tail degradation, and
$P_{80}-P_{50}$ characterizes upper-tail improvement. Because the
lower-tail term is assigned greater weight than the upper-tail term
under the default setting ($\beta=0.50>\gamma=0.25$), the aggregation
reflects the practical preference for avoiding unstable degradation
rather than rewarding occasional unusually strong outcomes.

\subsection{Interpretation and Limitations}

The values $\beta=0.50$ and $\gamma=0.25$ should therefore not be
interpreted as universally optimal parameters. Their appropriate values
can depend on the deployment objective. A safety-critical application,
for example, may reasonably assign a larger penalty to lower-tail
behavior, whereas an application that values occasional high-quality
responses may place greater weight on the upper tail.

For this reason, the proposed framework does not rely on percentile
scoring as the sole checkpoint-selection criterion. We use it as one
robust aggregation mechanism within Stage I and evaluate its sensitivity
over a range of parameter values. Finalists are subsequently examined
through listwise and pairwise comparison, and unresolved cases are
preserved as near-ties rather than being decided solely by the choice
of $\beta$ and $\gamma$.

\end{document}